# Classification of Approaches and Challenges of Frequent Subgraphs Mining in Biological Networks

Mohammadreza Keyvanpour
dept. Computer Engineering
Alzahra University
Tehran, Iran
Keyvanpour@Alzahra.ac.ir

Fereshteh Azizani
dept. Computer Engineering
Islamic Azad University, Qazvin Branch
Qazvin, Iran
Fereshteh.Azizani@gmail.com

*Abstract*— **Understanding the structure and dynamics of biological networks is one of the important challenges in system biology. In addition, increasing amount of experimental data in biological networks necessitate the use of efficient methods to analyze these huge amounts of data. Such methods require to recognize common patterns to analyze data. As biological networks can be modeled by graphs, the problem of common patterns recognition is equivalent with frequent sub graph mining in a set of graphs. In this paper, at first the challenges of frequent subgrpahs mining in biological networks are introduced and the existing approaches are classified for each challenge. then the algorithms are analyzed on the basis of the type of the approach they apply for each of the challenges.**

## I. INTRODUCTION

Understanding the structure and dynamics of biological networks is one of the important challenges in system biology [1,2]. System biology is a branch of bioinformatics and it aims to understand the type of properties of biological systems (molecule, cell, texture and organism) that not exist in single elements of these systems, but arise from their collective interaction [1]. The importance of this understanding in one hand and the increasing amount of experimental data in biological networks domain on the other hand, necessitate the use of efficient methods to analyze efficiently these amounts of data.

The most important biological networks are including as Protein interaction networks, Gene regulatory networks, and metabolic pathways. One of the main requirements in biological networks analysis is the recognition of the common parts in thems. Understanding the common parts result into the recognition of motifs, functional modules, relationships and interactions between sequences and patterns of gene regulation [3]. As , biological networks are mostly modeled with graphs in which nodes correspond to biomolecules and edges correspond to the interactions between them, it can be said that understanding the common parts is equivalent with frequent sub graph mining in a set of graphs. Frequent subgraphs mining is the main issue in graph mining domain. Graph mining or data mining in graphs is one of the important issues that are raised in data mining by the increasing amount of using graphs in modeling the complicated structures [4]. The existing definitions for data mining is also true for graph mining except that the type of the data on which graph mining is working are graphs.

Since then different algorithms are presented for frequent subgraphs mining in biological networks. This paper presents a general view about frequent sub-graph mining algorithms in biological networks. The algorithms are classified based on how they solve the challenges of frequent sub graph mining. The rest of the paper is organaized as follows. Section 2 defines the frequent graph mining. Section 3 introduces the different kinds of challenges in frequent subgraphs mining. Section 4 presents the approaches introduced for the solution of the challengers. Section 5 deals with the comparative analysis of the algorithms and section 5 present the conclusion.

## II. FREQUENT SUBGRAPHS MINING

The frequent subgraph is the one that occur frequently in the graph database**.** Graph database is special kinds of database that is mostly including a single large graph or some multiple small graphs. But in the area of biological networks, database is consisting of some large graph [3]. Indeed this issue can be explained exactly as the followings:

If D is the entry database, the frequent sub graph mining aims to mine graphs with more support value in comparison with predetermined threshold. The graph support $G_S$ is denoted by $sup(G_s)$ and is given as

$$\sup(G_s) = \frac{\text{The number of graphs of D included } G_S}{\text{The total number of graphs in D}} \quad (1)$$

## III. CHALLENGES

The investigation of the total presented methods for frequent sub graph mining reveals 3 challenges in this process. The challenges are introduced in this section.

### A. First Challenge;The Great Amount of Mined Subgraphs

Applying frequent subgraph mining algorithms on a set of graphs cause that all the subgraphs occurred more than a special threshold are discovered by algorithm [5]. These patterns are very much and these amounts of patterns are dependent upon the data property and defined threshold.

The considerable amount of patterns increase the operation time, make the selection of valuable patterns more difficult and reduce the scalability. So, there should be some approaches limiting search space and just mine frequent subgraphs with special conditions.

*B. Second Challenge; Sub Graph Isomorphism*

To determine the graph frequency number it is necessary to mine isomorphic graphs in D. If two sub graphs are similar in terms of connectivity, they are *isomorphic*. In other words, two graphs $G = (V_1, E_1)$ and $G' = (V_2, E_2)$ are *isomorphic* if there is a mapping from $V_1$ to $V_2$ such that each edge in $E_1$ is mapped to a single edge in $E_2$ and vice-versa. In labeled graphs, the labels should be added to the mapping [4]. The isomorphism review is a NP-complete issue and it is costly especially for large graphs [6].

*C. Third Challenge;Subgraphs Connectivity*

Most of the existing algorithms for frequent graph mining attempt to improve frequent itemsets mining algorithms for this case. Frequent itemsets mining begin with frequent items [7]. Frequent items are the ones the frequency number in database is higher than a special threshold. In first stage frequent 1-itemsets (frequent itemsets with one item) with are mined. frequent item is added to frequent 1-itemsets to create 2-itemsets. Since the resulting 2-itemsets are not frequent anymore and they are called candidate set or in short candidate. Then the frequency of the candidate or the resulting candidates is investigated and the frequent ones are returned as the next stage entry. This process is continuing till the algorithm time is achieving a predetermined threshold or all the frequent itemsets are mined. In graphs, it is necessary to keep graph connectivity in each stage.

IV. IDIFICATION AND CLASSIFICATION OF THE APPROACHES TO SOLVE THE CHALLENGES

The over mentioned challenges are common in all the graph mining functional areas. But as the database is consisting of a great amount of sparse graphs in biological networks area, the challenges be more intense. The main reason is that as the graphs are increasing and the size is bigger, the more subgraphs should be analyzed. The increase in the number of subgraphs increase the number of the discovered sub graphs, the isomorphism tests and the necessary connectivity tests. so, the existing approaches for frequent graph mining is inefficient for the biological networks.

This issue reveals the necessity of designing an efficient algorithm for biological networks in terms of time and memory. Fig. 1 shows the approaches to solve each one of the challenges. In the following section these approaches are introduced in brief. Indeed the main purpose here is the introduction of the existing approaches idea and the detailed explanation of them is not the required in this part.

*A. The Approachs to Solve the First Challenge*

The first approach to solve the problem is that only maximal frequent sub graphs are mined [8,9]. A graph $G'$ is said to be a maximal frequent subgraph if it satisfy the following conditions:

1) $G'$ is a frequent subgraph
2) $\nexists$ freduent subgraph $G''$ distinct frome $G'$ such that $G' \subseteq G'' \subseteq G$

As the number of the maximal subgraphs is less than the number of frequent subgraphs, it significantly reduces the total number of the mined subgraphs. Mining maximal subgraphs fulfills the requirement of biological networks. *Mule* [8] is the first algorithm using this idea that improved *Apriori* algorithm [10] for mining the biological networks such that by depth first search only maximal connected subgraphs are mined. H. Zantema et al' approach (*ZWB*) [9] is the second algorithm that only mine the maximal subgraphs to reduce the number of subgraphs.

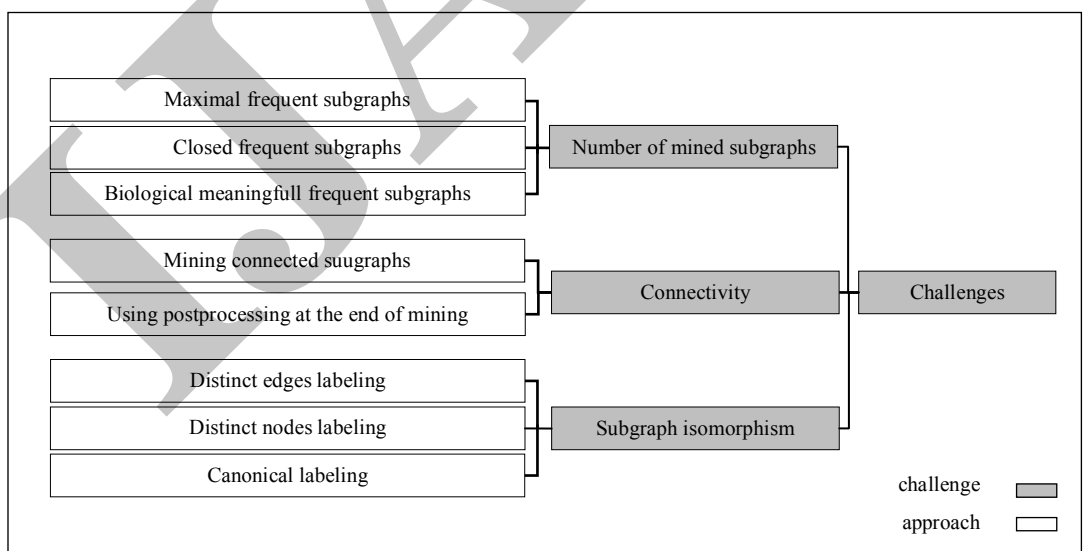

Figure 1. The approaches to solve the problems of frequent subgraph mining in biological networs

This algorithm like the previous algorithm by considering each graph as a set of its edges, change the issue of maximal frequent subgraphs mining to maximal frequent itemsets mining. The difference between this algorithm and mule is that it doesn't improve the maximal itemsets mining for maximal frequent graphs mining but use it for frequent subgraphs mining.

The second approach is the closed frequent subgraphs mining. A graph $G'$ is said to be a closed frequent subgraph if it satisfy the following conditions:

1) $G'$ is a frequent subgraph
2) $\nexists$ freduent subgraph $G''$ distinct frome $G'$ such that $G' \subseteq G'' \subseteq G$ and $\sup(G'') = \sup(G')$

*MAXFP* [11] is one of the algorithms that mine only closed frequent subgraphs in biological networks. This algorithm is based on a satisfaction model suitable for biological networks.

One of the other approaches is to mine biological meaningful subgraphs [12,13]. This idea was proposed as most of the mined subgraphs are not meaningful biologically and they can be ignored. For example, if we assume the network database in Fig. 2, frequent sub graph mining algorithms exploit the graphs including nodes *c, f, h, d ,g* and *e* But biologically, it is better to divide this graph into two modules with nodes *c, f, h, e* and *e, d, h, g*. Because these two modules have different occurrences in this database. In [12] *Condense* algorithm is presented to mine the subgraphs equivalent to biological modules. The other algorithms to mine biological meaningful subgraphs are the algorithm in [13] that in the rest of this paper is called *DGZY*. Each motif in Protein interaction networks and Gene regulatory networks is consisting of one or more Hamiltonian subghraphs. So Hamiltonian cycles are crucial for their biological performance. This algorithm using this idea, only detects subgraphs with Hamiltonian cycles and in this way the total number of the mined sub graphs is reduced.

*B. The Approaches to Solve the Second Challenge*

The first approach to solve the problem of isomorphism in graphs is the application of canonical labeling for each subgraph [14]. In this method by using nodes and edges labels, a distinct code is dedicated to each subgraph. This code is called graph standard label. Thus, instead of investigating the similarity of two graphs, it is adequate to ensure that whether two graphs have similar canonical label or not. But the computational complexy of the canonical labeling is also exponential in the worst case and they are not suitable due to the magnitude of biological networks.

The main source of isomorphism in mining of frequent subgraphs in labeled graphs is the repetition of the nodes lables. The second class of the approaches by this idea model biological network in a way that each node is having a distinct label. Any graph in which each node having a distinct label are called *relational* graph [15]. These approaches are based on the fact that the biological networks modeling with *relational* graph fulfill totally the research requirements. Thus, a graph is recognized with its edges and there is no need for ensuring about the isomorphism. The algorithms of this type are divided into two general groups.

The main difference of these two groups is the method they use to model the biological networks. As it was said before, biological networks are mostly modeled with graphs in which nodes correspond to biomolecules and edges correspond to the interactions between them. The first group algorithms consider one node in the graph for all the biomolecules with similar label in the related biological network [8,10,11]. There is edge between two nodes if two bimolecular interact with each other. The second group algorithms believe that this kind of display sometimes make to lose the data. For example, in some cases that two biomolecules have different interactions with different types, For all the interactions in the corresponding graph just one edge is considered. So, the type of the interaction between the biomolecules is not taken into consideration. Thus, although this kind of display meet the demands of most of the applications, it is necessary in some of the applications to save the data. This idea was first applied for Metabolic pathways in [16]. In this method (*LBH*) Metabolic networks are modeled with graphs in which for each interaction in biological network is considered a node in the corresponding graphs and all the data is saved in the graph.

*C. The Approaches to Solve the Third Challenge*

To test connectivity there are two approaches. First frequent itemsets mining algorithms are improved in a way that connectivity issue is also considered in them [9]. Second, regarding the graphs as the set of their edges, exactly frequent itemsets mining algorithms are used and finally connected sub graphs are created by post-processing [10,12].

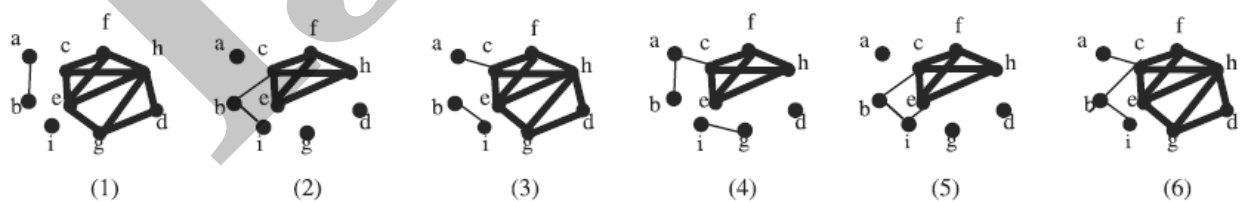

Figure 2. Database including 6 graphs with similar nodes and different edges [13]

The comparison of the results of the two groups of approaches shows the better performance of the second group in comparison with the first group. The main reason is that the number of the necessary connectivity tests at the end of the mining process is less than their total number during mining. Indeed, at the end of the process just the connectivity of the frequent sets is reviewed while, if the they are investigated during the mining, all the sets should be reviewed even if they are not recurring.

## V. THE ANALYSIS OF THE ALGORITHMS

Table 1 analyzes the mentioned algorithms on the basis of the type of approach they apply for each of the above challenges. All these algorithms claim that their recommended approach is applicable in all the different kinds of biological networks but they present the results of Algorithm on one type of network. Column *BN* of the table show the type of the network the related algorithm was tested on.

## VI. CONCLUION

The paper analyzed the frequent subgraphs mining algorithms in the biological networks domain. At first the related challenges were introduced and the existing approaches were presented for each challenge. then the algorithms were analyzed on the basis of the type of the approach they would apply for each of the challenges. The results show that most of the existing algorithms have been designed for Metabolism pathways. The issue reveals the necessity to create suitable algorithms for the other kinds of biological networks. The above classification can help to create such approaches.

TABLE I.    COMPARISON OF SUBGRAPHS MINING ALGORITHMS CHARACTERISTICS

| Algorithm | Year | GDB topology | subgraphs type | Graph isomorphism | connectivity | BN |
|---|---|---|---|---|---|---|
| **Mule** | 2004 | Directed | Maximal | Distinct nodes label | In mining | metabolic pathways |
| **CODENSE** | 2005 | Undirected | Coherent dense | Distinct nodes label | In mining | protein interaction network, genetic interaction network and co-expression networks |
| *MaxFP* | 2008 | Directed | Closed | Distinct nodes label | After mining | metabolic pathways |
| *ZWB* | 2008 | Directed | Maximal | Distinct nodes label | After mining | metabolic pathways |
| **LBH** | 2009 | Directed | Maximal | Distinct edges label | After mining | metabolic pathways |
| **DGZY** | 2007 | directed | Hamiltonian | adjacency matrix | In mining | protein interaction network, genetic interaction network |